\documentclass[a4paper,11pt]{article}
\usepackage{cite}
\usepackage{latexsym,amssymb,enumerate,amsmath,epsfig,amsthm,dsfont,bm}
\usepackage[margin=1in]{geometry}
\usepackage{setspace,color}
\usepackage{tikz}
\usepackage{floatrow}
\usepackage{graphicx,subfigure}
\usepackage[ruled]{algorithm2e}
\usepackage{epstopdf}
\usepackage[multidot]{grffile}
\usepackage{comment}
\usepackage{multirow}

\makeatletter
\newcommand\figcaption{\def\@captype{figure}\caption}
\newcommand\tabcaption{\def\@captype{table}\caption}
\makeatother

\theoremstyle{plain}

\theoremstyle{definition}

\newtheorem{?}[Th]{Problem}

\newcommand{\bc}{\mathbf{c}}

\begin{document}
	
\title{Underwater Image Color Correction by Complementary Adaptation}
\author{Yuchen He
}
\date{}
\maketitle

\begin{abstract}
	In this paper, we propose a novel approach for underwater image color correction based on a Tikhonov type optimization model in the CIELAB color space. It presents a new variational interpretation of the complementary adaptation theory in psychophysics, which establishes the connection between colorimetric notions and color constancy of the human visual system (HVS). Understood as a long-term adaptive process, our method  effectively removes the underwater color cast and yields a balanced color distribution. For visualization purposes, we enhance the image contrast by properly rescaling both lightness and chroma without trespassing the  CIELAB gamut.  The magnitude of the enhancement is hue-selective and image-based, thus our method is robust for different underwater imaging environments. To improve the uniformity of CIELAB, we include an approximate hue-linearization as the pre-processing and an inverse transform of the Helmholtz-Kohlrausch effect as the post-processing.  We analyze and validate the proposed model by various numerical experiments.  Based on image quality metrics designed for underwater conditions, we  compare  with some state-of-art approaches to show that the proposed method has consistently superior performances.
\end{abstract}

\section{Introduction}
%
%
%
%

\begin{figure}
	\centering
	\includegraphics[scale=0.15]{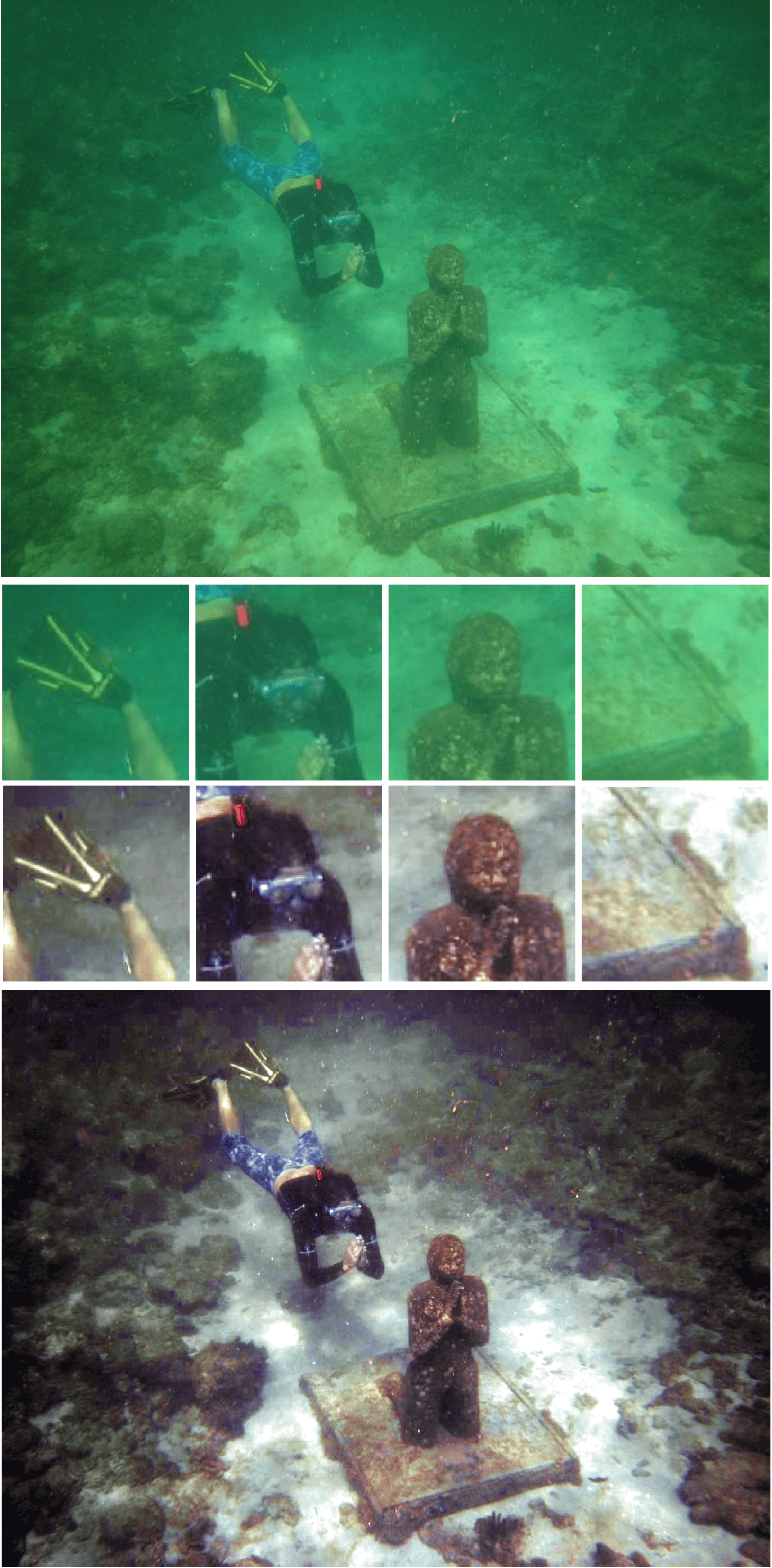}
	\caption{(Top) Underwater image with heavy green cast. (Bottom) Result of the proposed method. In the middle, several zoomed-in regions are displayed for comparison. The resulted image has enhanced contrast, balanced colors, and many image contents, e.g., the patterns on the swimming shorts, are more recognizable. In this paper, all the underwater images are from the benchmark data set~\cite{li2019underwater}. }\label{fig_demo}
\end{figure}

Water absorbs light similarly to an optical filter but with higher variations and complexities~\cite{zhang2001underwater}. Depending on the dissolved or suspended substances, a liquid medium modifies the spectral power distribution of the transmitted light, such that a strong bluish or greenish color cast dominates the acquired underwater image, e.g., Figure~\ref{fig_demo} (Top). Typically, underwater images have insufficient contrast and unbalanced color distribution~\cite{zhang2001underwater,yang2015underwater,wen2013single,peng2017underwater,berman2017diving}, hence many image contents, such as patterns and textures are hardly recognizable for human observers. An effective color correction method is needed to recover and enhance these details.


We can understand this task as reversing the process of image formation. The complex factors determining the irradiance on an imaging sensor are often simplified by the Koschmieder model~\cite{koschmieder1924theorie}. It expresses the image colors as a convex combination of the unattenuated objects' colors and the veiling light via a  scalar transmission map.  The veiling light is approximated by various types of dark-channel priors~\cite{yang2015underwater,wen2013single,peng2017underwater}, and the estimation of the transmission map is converted to depth computation based on the Beer-Lambert law~\cite{swinehart1962beer}. Hence, for any combination of a veiling light and a transmission map, the color-corrected image is uniquely determined. These methods are  sensitive to the identified veiling lights such that small perturbations on the estimated RGB values of the background trigger visually significant results~\cite{peng2017underwater}. More sophisticated models along this direction consider the Jerlov's water types~\cite{jerlov1968irradiance,berman2017diving} to improve the stability.

Essentially, we are finding a color distribution on the image domain that is favorable for a human observer.  Different from models of physics, many methods in the literature adjust the image colors based on principles of the human visual system (HVS). One of the  important properties of HVS is color constancy:  the appearance of the color of an object remains approximately stable under varying illuminations~\cite{10.2307/24949866}. This chromatic adaptation for instance allows an observer to recognize the brown statue and the blue shorts in Figure~\ref{fig_demo} (Top), even though the image is dominated by a heavy green cast. Theories~\cite{finlayson2000improving,moore1991real,finlayson1994spectral} have been proposed to explain the underlying mechanism from various perspectives including the well-known Retinex theory by Land~\cite{land1977retinex}. It is argued that HVS perceives a scene based on local variation of image lightness rather than an absolute lightness, and this theory induces a huge class of algorithmic interpretations of the adaptation process applied in computer vision, e.g., Multiscale Retinex~\cite{rahman1996multi,ipol.2014.107}, random-spray Retinex~\cite{provenzi2006random},    non-local Retinex~\cite{zosso2015non} and many others~\cite{morel2009fast,kimmel2003variational,morel2010pde}.

\begin{figure*}[ht]
	\centering
	\includegraphics[width=0.8\textwidth]{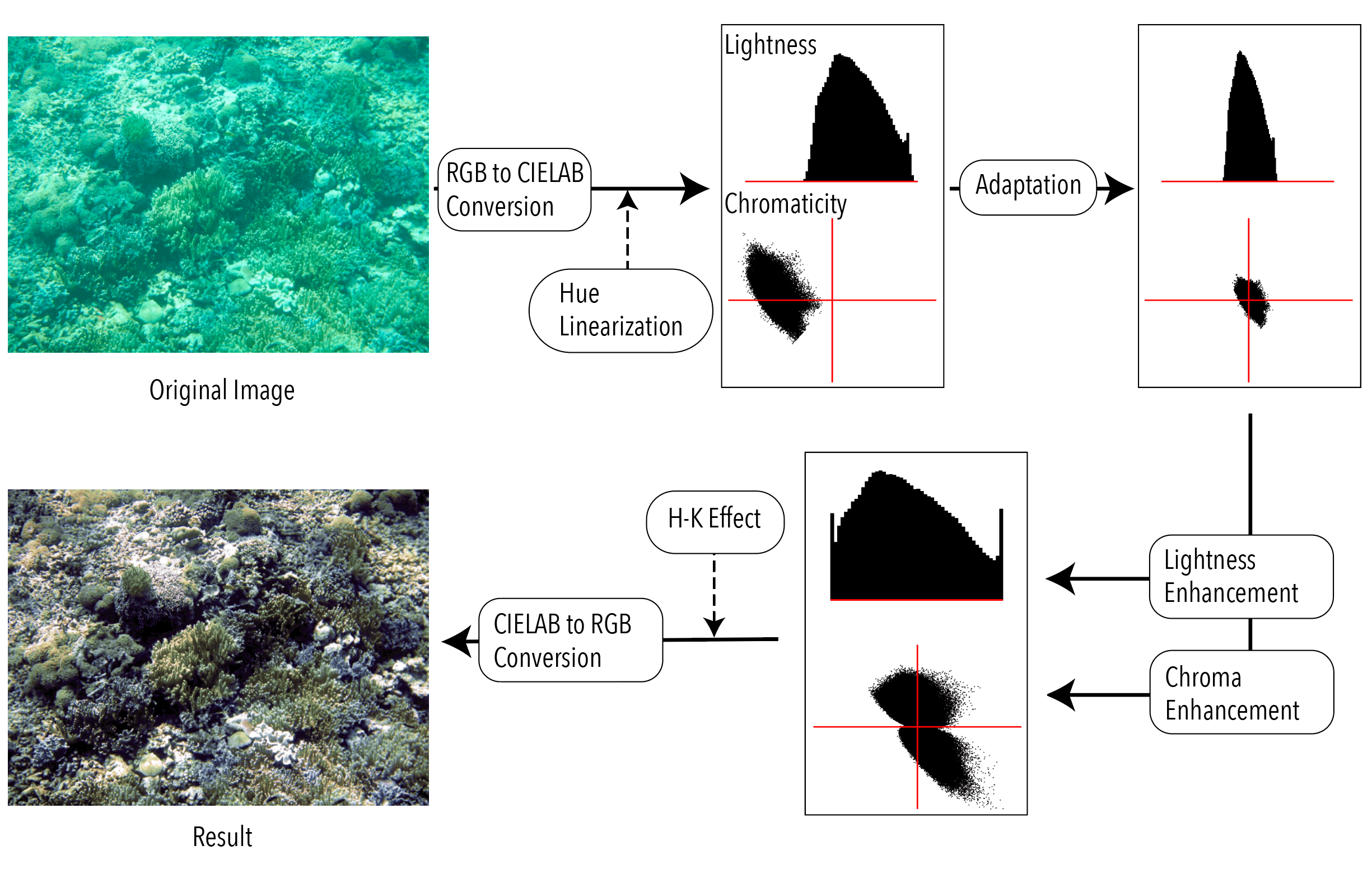}
	\caption{Pipeline of the proposed method. The result shown here uses $\eta=10,\beta=1/3$ as the model parameters.}\label{fig_outline}
\end{figure*}

In this paper, we propose a novel  approach for underwater image color correction which converts Figure~\ref{fig_demo}~(Top) to (Bottom), whose color distribution is more balanced and compatible with HVS. Instead  of the Retinex theory, we present a new mathematical interpretation for the Complementary Adaptation Theory (CAT)  first formulated by Gibson~\cite{gibson1937adaptation} in 1937. The key principle is  that, the quality of a constantly applied stimulus will be temporarily shifted towards the corresponding complementary quality, thus resulting in a neutral state.  This applies not only to HVS, but also to other bilateral sensory processes, e.g., temperature perception. Both Retinex theory and CAT emphasize the importance of relative levels over absolute levels of sensation, yet they are fundamentally different in the following aspects.
\begin{itemize}
	\item \textit{Mechanism}: The Retinex theory ascribes the color constancy to HVS's ability of estimating the reflectance independent from the illumination, while CAT describes the color constancy as a result from  the neutralization of the illumination by a negative sensory  process. 
	\item \textit{Role of illuminating color}: In Retinex theory, illumination is treated as unknown  and its color can be derived after identifying the reflectance; whereas in CAT, the illuminating light  determines the direction and magnitude of the adaptation process. 
	\item \textit{Adapting time}:  The Retinex theory was supported by experiments with short-time adapting; in contrast, recent experiments show that the chromatic neutralization predicted by CAT occurs after multiple days~\cite{belmore2011very,tregillus2019long}. 
\end{itemize}
Our method captures these features of CAT and produces a color distribution complying with the long-term chromatic adaptation. Figure~\ref{fig_outline} shows the outline of our method.

The idea is that, we utilize the complementary pairs of the locally approximated color cast to modify the image colors, such that any image colors similar to the color cast are muted, while the others keep their differences relative to the color cast.  In other words, we shift the reference color from the chromatic color cast, typically blue and green, to a neutral gray. The resulted color distribution has softer contrast and lower saturation due to the long-term adaptation. Hence, for visualization purposes, we enhance the image while preserving the adapted hues. 

In particular, we consider the CIELAB color space and formulate the CAT adaptation process as a Tikhonov-type optimization problem~\cite{tikhonov1963solution}. As a metric space, CIELAB is a subspace of the three dimensional Euclidean space, where the distance between any two colors measures their perceptive difference. Using the CIELAB color difference metric, our optimization model consists of a fidelity term and a regularization characterizing the behavior of CAT adaptation. Then we enhance the adapted color distribution for visualization purposes. We also address some technical issues about CIELAB to improve its uniformity. These modifications are kept minimal so that problematic behaviors are effectively adjusted, and  high efficiency is achieved.

To summarize, our contributions in this paper are:
\begin{enumerate}
	\item We propose a novel approach for underwater image color correction  using a Tikhonov-type optimization model in CIELAB color space.
	\item  We present a new mathematical interpretation of the complementary adaptation theory by Gibson for the color constancy of HVS. 
	\item We design a simple procedure to effectively improve the uniformity of CIELAB.
\end{enumerate}

We organize this paper as follows.  In Section~\ref{sec_proposed}, we present in detail our proposed method. In Section~\ref{sec_result}, we conduct various numerical experiments to analyze the model's behaviors and quantitatively compare our method with some state-of-art techniques based on underwater-specific image quality metrics. We conclude the paper in Section~\ref{sec_conclude} and include some technical details in the Appendix.

\section{Proposed Method}\label{sec_proposed}
\subsection{Notations}
Before stating our model, we fix some notations. We denote an arbitrary color by $\mathbf{c}$, which is specified in the CIELAB color space by its lightness $L^*$, red-green value: $a^*$, and  yellow-blue value: $b^*$, i.e., $\mathbf{c}=(L^*,a^*,b^*)$. In particular, the range for $L^*$ is $[0,100]$, where $L^*=0$ yields black and $L^*=100$ yields diffuse white. The $a^*$-$b^*$ section specifies the chromaticity. A positive value of $a^*$ indicates red while a negative value gives green. A positive value of $b^*$ indicates yellow and a negative value gives blue.   From CIELAB, one can derive
\begin{align}
	&\text{Chroma:}\quad C^*=\sqrt{(a^*)^2+(b^*)^2}\;,\label{eq_chroma}
\end{align}
which measures the relative saturation of $\mathbf{c}$, and
\begin{align}
	&\text{Hue angle:}\quad h^\circ=\text{atan2}(b^*,a^*)\;,\label{eq_hue}
\end{align}
which defines the hue of $\mathbf{c}$. 

Given a pair of colors $\mathbf{c}_i =(L_i^*,a_i^*,b_i^*)$, $i=1,2$, the CIELAB color difference between them is computed by
\begin{align} 
	\Delta E^*(\mathbf{c}_1,\mathbf{c}_2)= \sqrt{(L_1^*-L_2^*)^2+(a_1^*-a_2^*)^2+(b_1^*-b_2^*)^2}\;,\label{eq_colordiff}
\end{align}
which is simply the Euclidean distance between the CIELAB coordinates of the colors to be compared. This formula suggests that CIELAB color space is designed to be uniform. 

The complementary color of $\mathbf{c}$ in CIELAB is denoted by $\mathbf{c}^-$, which is computed by
\begin{align}
	\mathbf{c}^- = (100-L^*,-a^*,-b^*)\;.
\end{align}

On a rectangular image domain $\Omega=[0,W]\times [0,H]\subset\mathbb{R}^2$, $W,H>0$, we define a color image, or a color distribution  as a mapping $\mathbf{c}$ from $\Omega$ to the CIELAB color space 
\begin{align}
	\mathbf{c}(x,y)=(L^*(x,y),a^*(x,y),b^*(x,y))\;,~(x,y)\in\Omega\;.\label{eq_cielabimage}
\end{align}

For a triplet $\sigma = (\sigma_1,\sigma_2,\sigma_3)\in\mathbb{R}^3$ with positive entries, we define the component-wise Gaussian convolution $\mathcal{G}_\sigma$ applied on a color distribution $\mathbf{c}$ as 
\begin{align}
	&\mathcal{G}_\sigma*\mathbf{c}(x,y) =
(\mathcal{G}_{\sigma_1}*L^*(x,y),\mathcal{G}_{\sigma_2}*a^*(x,y),\mathcal{G}_{\sigma_3}*b^*(x,y))\;,
\end{align}
where each component is the ordinary Gaussian convolution with intensity specified by $\sigma_i$, $i=1,2,3$. We take mirror reflect for computing the convolved values near the image boundary.

\subsection{Complementary Adaptation Model in CIELAB}
Given a  color distribution $\bc_0=(L_0^*,a_0^*,b_0^*)$ over $\Omega$, we  propose the \textbf{Complementary Adaptation Model} by defining the adapted color at $(x,y)\in\Omega$ as the minimizer of the following optimization problem
\begin{align}
	\bc_{\text{adapt}}(x,y)&= \arg\min_{\bc\in\mathbb{R}^3}\left(\Delta E^*(\bc(x,y),\bc_0(x,y))\right)^2+\lambda\left(  \Delta E^*(\bc(x,y),(\mathcal{G}_\sigma*\bc_0(x,y))^-)\right)^2\;,\label{eq_opt_model}
\end{align}
where  $\lambda>0$ is a weight parameter, and $\sigma = (\sigma_{L^*},\sigma_{a^*}, \sigma_{b^*})$ such that $\sigma_{a^*}=\sigma_{b^*}$ and $\sigma_{L^*}=n\sigma_{a^*}$ for some $n>1$. This is a Tikhonov-type optimization problem consisting of two terms. The first term measures the difference between the given color distribution $\bc_0$ and the adapted color $\bc_{\text{adapt}}$, thus it imposes the fidelity condition. The second term models the effect of CAT adaptation process, which acts as a regularization. The proposed color distribution $\bc_{\text{adapt}}$ is a balance between the original image  and the  complementary of the estimated color cast $\mathcal{G}_\sigma*\mathcal{C}$. In this paper, we fix $\sigma_{a^*}=\sigma_{b^*}=\sigma_0 := 0.25(\max(W,H)/2-1)$ so that the size of the filter is roughly $\max(W,H)/2$, $n=3$, and $\lambda=1$. 

Thanks to the simple formula~\eqref{eq_colordiff} for computing the color difference, \eqref{eq_opt_model} has a unique global minimizer obtained by calculus:
\begin{align}
	\bc_{\text{adapt}}(x,y) = (L_{\text{adapt}}^*(x,y),a_{\text{adapt}}^*(x,y),b_{\text{adapt}}^*(x,y))\;,\label{eq_adapted}
\end{align}
where
\begin{align}
	\begin{cases}
		L_{\text{adapt}}^*(x,y) = \left(L_0^*(x,y)+(100-\mathcal{G}_{3\sigma_0}*L_0^*(x,y))\right)/2\\
		a_{\text{adapt}}^*(x,y) = \left(a_0^*(x,y)-\mathcal{G}_{\sigma_0}*a_0^*(x,y))\right)/2\\
		b_{\text{adapt}}^*(x,y) = \left(b_0^*(x,y)-\mathcal{G}_{\sigma_0}*b_0^*(x,y))\right)/2
	\end{cases}\label{eq_adapt_formula}
\end{align}
This formulation shows that the adapted color $\bc_{\text{adapt}}(x,y)$ is the midpoint of the image color $\bc_0(x,y)$ and the complementary pair of the estimated color cast at $(x,y)$ in the CIELAB space. 

Some remarks are needed for the proposed model: 

\begin{enumerate}
	
	\item\textit{Locality principle}: The adaptation is spatially dependent~\cite{mccann1987local}. Notice that in the second term of the model, the complementary operator is applied to the  Gaussian filtered  color distribution instead of the original $\mathbf{c}_0$. The locality of the adaptation is adjusted by the parameter $\sigma_0$. A greater value of $\sigma_0$ implies a larger field of adaptation and the estimated color cast is spatially more uniform. In contrast, a smaller value of $\sigma_0$ induces a more focused adaptation and the estimated color cast is more variant. Consequently, the neutralization effect is stronger when $\sigma_0$ is small; when $\sigma_0\to 0 $, the adapted color distribution becomes uniformly neutral gray. 
	
	\item\textit{CIELAB gamut consideration}: In practice, the 3-tuple $\bc_{\text{adapt}}$ computed by~\eqref{eq_adapt_formula} stay inside the CIELAB gamut. There are two reasons to support this statement. First, the adapted lightness $L^*_{\text{adapt}}$ concentrates around $50$, where the chromaticity section of the CIELAB gamut has the most extended domain (Figure~\ref{fig_gamut}~(a) and (b)). Second, the dominating color casts in underwater images are mostly blue or green. Observe that the CIELAB gamut (Figure~\ref{fig_gamut}~(c)) corresponding to the green-blue region only has limited expansion, hence both $\mathcal{G}_{\sigma_0}*a_0^*$ and $\mathcal{G}_{\sigma_0}*b_0^*$ are relatively small. Consequently, the triangle spanned by $(a_0^*,b^*_0)$ and $(\mathcal{G}_{\sigma_0}*a_0^*,\mathcal{G}_{\sigma_0}*b_0^*)$ is most likely  contained in the chromaticity domain at the lightness $L^*_{\text{adapt}}$. For robustness, in case $\bc_{\text{adapt}}(x,y)$ for some $(x,y)$ falls outside the CIELAB gamut, we keep its adapted lightness and hue angle while shrinking its chroma to the corresponding  maximal chroma. Other possible solutions can be found in~\cite{azetsu2019hue} and~\cite{ueda2017lightness}.
	
	\item\textit{Long-term adaptation}: The proposed color distribution $\bc_{\text{adapt}}$ is based on the neutralization of dominant colors, which is a long-term chromatic adaptation that can take multiple  days~\cite{belmore2011very,tregillus2019long}. This is different from the daily experience where the time of adaptation ranges from seconds to a few minutes~\cite{shevell2001time}.  Similarly to the experimental setting~\cite{belmore2011very}, the Gaussian filtered color distribution can be considered as colored lenses, and the long-term adaptation behavior is modeled by the regularization term. Hence, our model predicts the perceived colors when the observer wears the lenses for a long time and the dominant colors are neutralized by their complementary pairs, respectively. 
	
	\item\textit{Connection to other works}: The proposed model~\eqref{eq_opt_model} also provides a variational substitute for the well-known Gray World (GW) assumption~\cite{buchsbaum1980spatial},  which  is a key component in many methods in the literature, e.g., ACE~\cite{bertalmio2007perceptual}. In~\cite{bertalmio2007perceptual}, Bertalm\'{i}o et al. connect ACE to the Wilson-Cowan equations~\cite{wilson1972excitatory} from computational neuroscience, where the GW assumption is used to set an absolute neutral state  such that only deviations from this level are considered meaningful. Noticing the drawback of using an absolute level, in~\cite{bertalmio2014image}, Bertalm\'{i}o proposes to replace it with a local average; however, by doing so, no color correction is in action.  Our model provides an elegant solution which maintains an effective color correction while avoiding an absolute reference. 
\end{enumerate}

\subsection{Robust Hue-preserving Image Enhancement}\label{sec_enhance}
The adapted color distribution $\bc_{\text{adapt}}$ represents  a long-term result  rarely achieved in common life experience. For visualization purpose, we enhance the lightness $L^*_
{\text{adapt}}$ and the chroma $C^*_{\text{adapt}}$~\eqref{eq_chroma} computed using $a_{\text{adapt}}^*$,  $b_{\text{adapt}}^*$, while preserving the adapted hue $h^\circ_{\text{adapt}}$~\eqref{eq_hue} where the dominant color cast has been neutralized.

We enhance the adapted lightness by a linear stretch. The enhanced lightness is denoted by $\widehat{L^*}_\text{adapt}$.  To keep the transform consistent with the CIELAB gamut, we rescale the chroma of the adapted colors by
\begin{align}
	C_1^*(x,y) &=\left(\frac{C^*_{\text{adapt}}(x,y)}{C_{\max}^*(L^*_{\text{adapt}}(x,y),h^\circ_{\text{adapt}}(x,y))}\right)\times C_{\max}^*(\widehat{L^*}_{\text{adapt}}(x,y),h^\circ_{\text{adapt}}(x,y))
	\;.\label{eq_rescale2}
\end{align}
which preserves the percentage of the relative saturation of $\bc_{\text{adapt}}$. Here $C^*_{\max}(L^*,h^\circ)$ denotes the maximal chroma in the CIELAB gamut when the lightness is $L^*$ and the hue angle is $h^\circ$, whose computation is detailed in the Appendix~\ref{sec_gamut_boundary}. Notice that the adapted hue angles are unchanged during this rescaling.

\begin{figure}
	\centering
	\includegraphics[width=3.1 in]{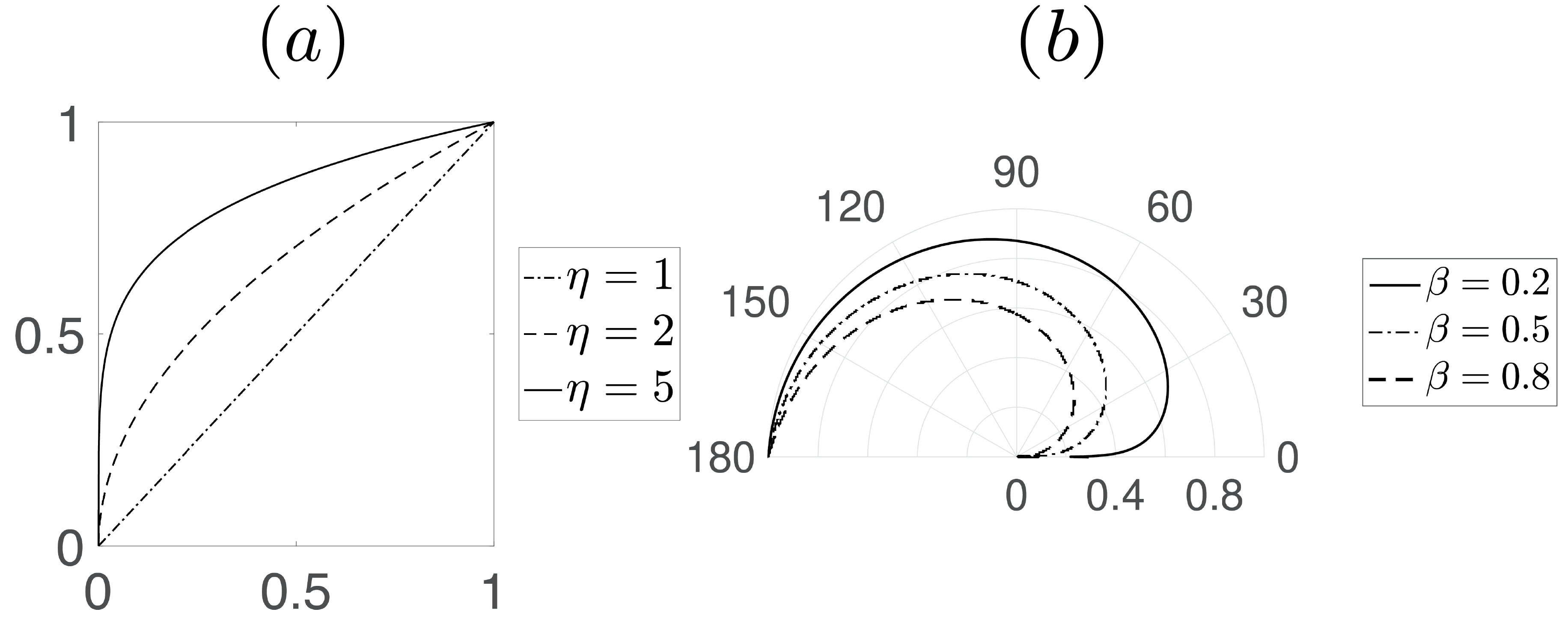}
	\caption{(a) Gamma function used for chroma enhancement~\eqref{eq_gamma_enhance} with varying values of $\eta$. (b) Robust factor~\eqref{eq_zeta} for suppressing the noisy hues with varying values of $\beta$. }\label{func_demo}
\end{figure}

For the fixed lightness $\widehat{L^*}_{\text{adapt}}$, we enhance the chroma of the newly obtained color distribution $(\widehat{L^*}_{\text{adapt}},a^*_1,b^*_1)$ by the following transformation
\begin{align}
	C_2^*(x,y) &= \left(\frac{C^*_1(x,y)}{C_{\max}^*(\widehat{L^*}_{\text{adapt}}(x,y),h^\circ_{\text{adapt}}(x,y))}\right)^{1/\eta}\times C_{\max}^*(\widehat{L^*}_{\text{adapt}}(x,y),h^\circ_{\text{adapt}}(x,y))\;.\label{eq_gamma_enhance}
\end{align}
We note that this is a gamma correction applied to the percentage of relative saturation. See Figure~\ref{func_demo} (a). Here, $\eta\geq 1$ is the enhancing parameter. When $\eta$ increases, a stronger enhancement is applied, and when $\eta=1$, the gamma function reduces to the  identity map. 

To improve stability,  for some $0<\beta\leq1$, we define 
\begin{align}
	&\text{Robust factor:}\quad F(\theta) = \left(\frac{\theta}{180^\circ}\right)^{\beta}\;,\label{eq_zeta}\\
	&\text{for any~}\theta\in [0,180^\circ]\;,\nonumber
\end{align}
whose behavior is shown in Figure~\ref{func_demo} (b). We propose the robust hue-preserving enhancement of the adapted color distribution $\bc_{\text{adapt}}$ by
\begin{align}
	\widehat{\bc}_{\text{adapt}}(x,y) = (\widehat{L^*}_{\text{adapt}}(x,y),\widehat{a^*}_{\text{adapt}}(x,y),\widehat{b^*}_{\text{adapt}}(x,y))\;,\label{eq_enhanced}
\end{align}
where 
\begin{align}
	\begin{cases}
		\widehat{a^*}_{\text{adapt}}(x,y)= F(\theta(x,y))a_2^*(x,y)\\
		\widehat{b^*}_{\text{adapt}}(x,y) =F(\theta(x,y))b_2^*(x,y)
	\end{cases}\;.\label{eq_adapt_formula2}
\end{align}
and $\theta(x,y)$ denotes the hue angle difference between the image color and the estimated color cast at $(x,y)$. Notice that by multiplying the robust factor, when $\theta(x,y)\approx0^\circ$, i.e., the hue angle difference between the image color and the estimated color cast is small, $\widehat{\bc}_{\text{adapt}}(x,y)$ becomes almost achromatic. As for image colors deviating from the estimated color cast at the same locations, the differences are emphasized.

\subsection{Improvement on the Uniformity of CIELAB }
The main purpose of the CIELAB  as an alternative to RGB is to quantify the perceptive color difference. This is only approximate due to the intrinsic complexity and non-linearity of the HVS. In  this work, we employ two simple modifications to achieve a better uniformity. 

\subsubsection{Adjustment in the Blue Region}\label{sec_pre}
As known to many researchers~\cite{mclaren1980cielab,ebner1998finding,braun1998color,moroney2003hypothesis},  the blue region of CIELAB, which roughly corresponds to the subset of colors with  hue angles ranging from $250^\circ$ to $300^\circ$, is not hue-linear. It means that, with the CIELAB lightness and hue angle fixed, increasing the CIELAB chroma yields a perceivable hue-shift.

In this work, we propose to address this technical problem by applying the following transform before solving for the adapted color distribution via~\eqref{eq_opt_model}:

\begin{align}
	&h^\circ_{\text{adjust}} = h^\circ -\mu^\circ\times\sqrt{\frac{(C^*)^m}{(C^*)^m+10^m}}\times \exp\left(-\left(\frac{h^\circ-275^\circ}{25^\circ}\right)^2\right)\;.\label{eq_adjust}
\end{align}
This formula is modified from~\cite{luo2001development}, which adjusts the hue angle by a product of three factors. The first factor $\mu^\circ$ denotes the maximal distorted hue angle. The second factor predicts the increase of the hue rotation from neutral, i.e., $C^*=0$ until around $C^*=10$ and remains constant in the high chroma region~\cite{luo2001development}. The last factor restricts the hue adjustment within the region $275^\circ<h^\circ<300^\circ$. We fix $\mu^\circ=45^\circ$ and choose
$m=7$ in this paper. For a more precise hue adjustment based on a look-up-table, we refer the readers to~\cite{braun1998color}. 

\begin{figure}
	\centering
	\includegraphics[width=3in]{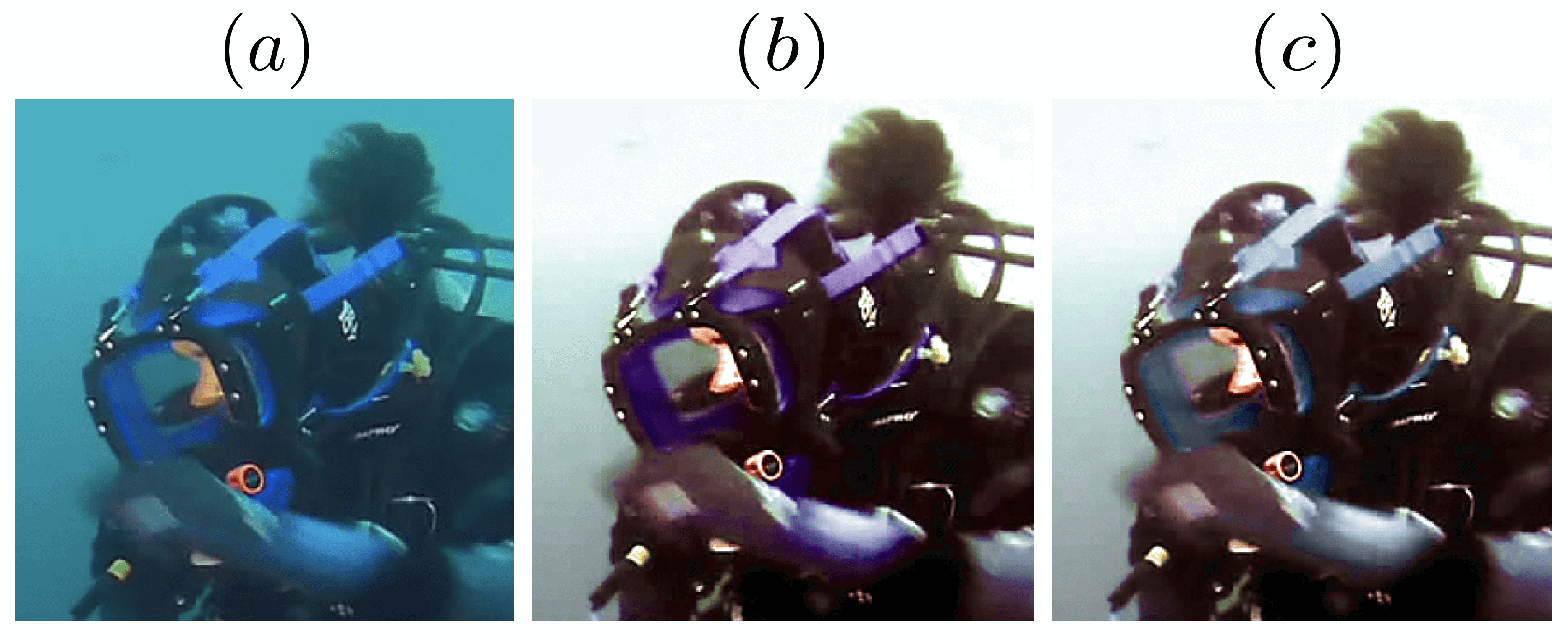}\\
	\caption{Pre-processing by hue angle adjustment in the blue region of CIELAB. (a) Part of an underwater image. (b) Proposed method without the pre-processing~\eqref{eq_adjust}. (c) Proposed method with the pre-processing. With the adjustment, the blueness on the strap is  preserved.}\label{fig_hue_dist}
\end{figure}

For underwater images, the pre-processing~\eqref{eq_adjust} is especially important, since the general color distributions concentrate around the blue region. In Figure~\ref{fig_hue_dist}, we apply the proposed method to an underwater image (a) without the hue adjustment~\eqref{eq_adjust} and the blue goggle straps turn into purple (b).  With the hue correction (c), we observe that the blueness is correctly preserved. Hence, including the  adjustment~\eqref{eq_adjust} as a pre-processing compensates for the distortion of the hue angle as we enhance the chroma.

\subsubsection{The Helmholtz-Kohlrausch Effect}\label{sec_post}
The Helmholtz-Kohlrausch (H-K) effect is a perceptual phenomenon where  the perceived lightness of a color with increasing saturation is brighter~\cite{donofrio2011helmholtz}. To enhance the chroma~\eqref{eq_adapt_formula2} while keeping the perceived  lightness unchanged,  we need to adjust $\widehat{L^*}_{\text{adapt}}$. For an arbitrary color $\bc = (L^*,a^*,b^*)$ in CIELAB, the perceived lightness $L^*_{\text{H-K}}$ when considering the H-K effect can be estimated by~\cite{fairchild1991predicting}
\begin{align}
	L^*_{\text{H-K}} = L^*+(2.5-0.025L^*)g(h^\circ)C^*\;,\label{eq_HK}
\end{align}
where
\begin{align}
	g(h^\circ) &= 0.116\times\Big|\sin\left(\frac{h^\circ-90^\circ}{2}\right)\Big|+0.085\;.\label{eq_g}\\
\end{align}
Hence, assuming that $\widehat{L^*}_\text{adapt}$ corresponds to the perceived lightness before the chroma enhancement, the associated CIELAB lightness after the chroma enhancement is computed by inverting~\eqref{eq_HK}, which gives
\begin{align}
	(\widehat{L^*}_{\text{adapt}})_{\text{adjust}} = \frac{\widehat{L^*}_{\text{adapt}}-2.5g(\widehat{h^\circ}_{\text{adapt}})\widehat{C^*}_{\text{adapt}}}{1-0.025g(\widehat{h^\circ}_{\text{adapt}})\widehat{C^*}_{\text{adapt}}}\;.\label{eq_HKadjust}
\end{align}
Similar improvement is also considered in~\cite{ueda2017lightness} for food image enhancement. 

\begin{figure}
	\centering
	\includegraphics[width=3.1 in]{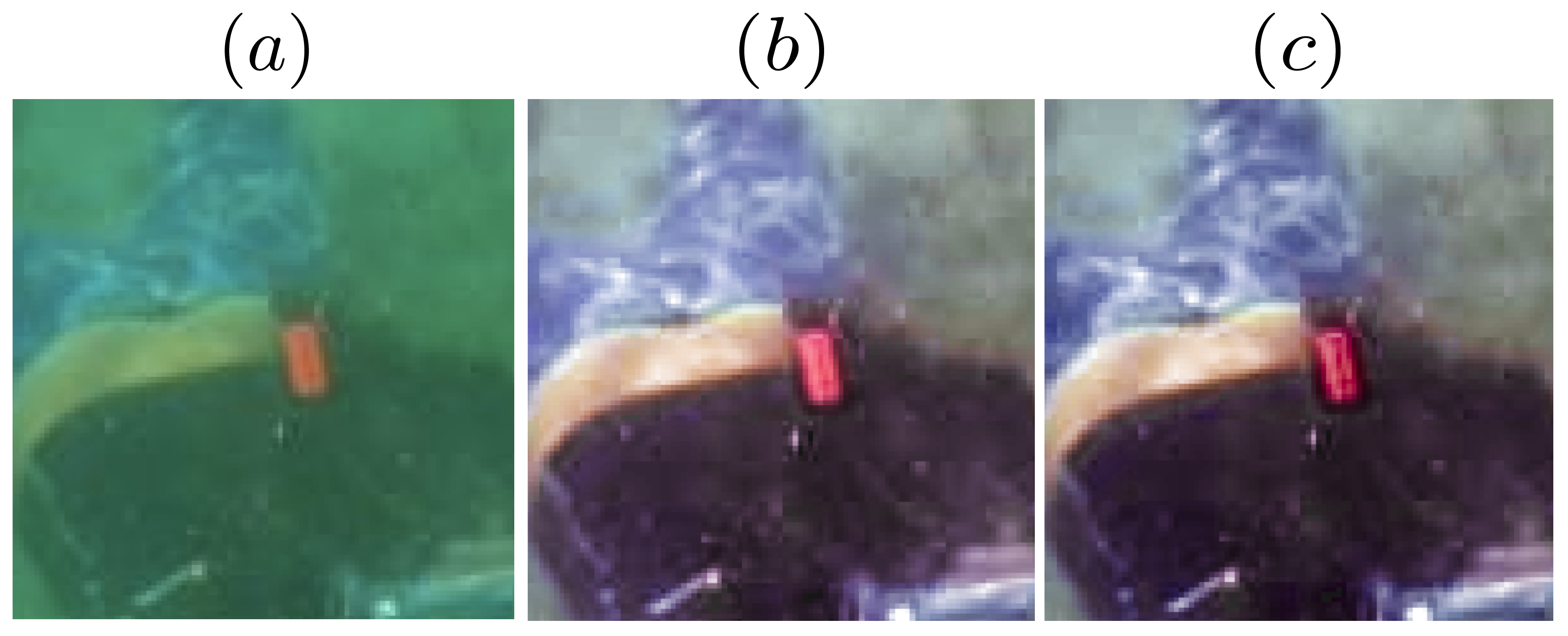}
	\caption{Post-processing considering the HK-effect. (a) Zoom-in of part of the underwater image in Figure~\ref{fig_demo}. (b) Proposed method without the post-processing~\eqref{eq_HK}. (c) Proposed method with the post-processing. Post-processing considering the HK-effect reduces over-exposure.}\label{fig_HK_effect}
\end{figure}

The post-processing in regard to the HK-effect~\eqref{eq_HK} addresses the over-exposure caused by the enhancing saturation.  In Figure~\ref{fig_HK_effect}, we focus on a zoomed-in region of an underwater image (a) and show the result without the post-processing (b) as well as the processed one (c). Comparing these results, we observe that when the HK-effect is considered, the image contrast is also improved. See the patterns on the shorts and the red tube. 

\section{Numerical Results}\label{sec_result}
\subsection{General Examples}

\begin{figure*}
	\centering
	\includegraphics[scale=0.18]{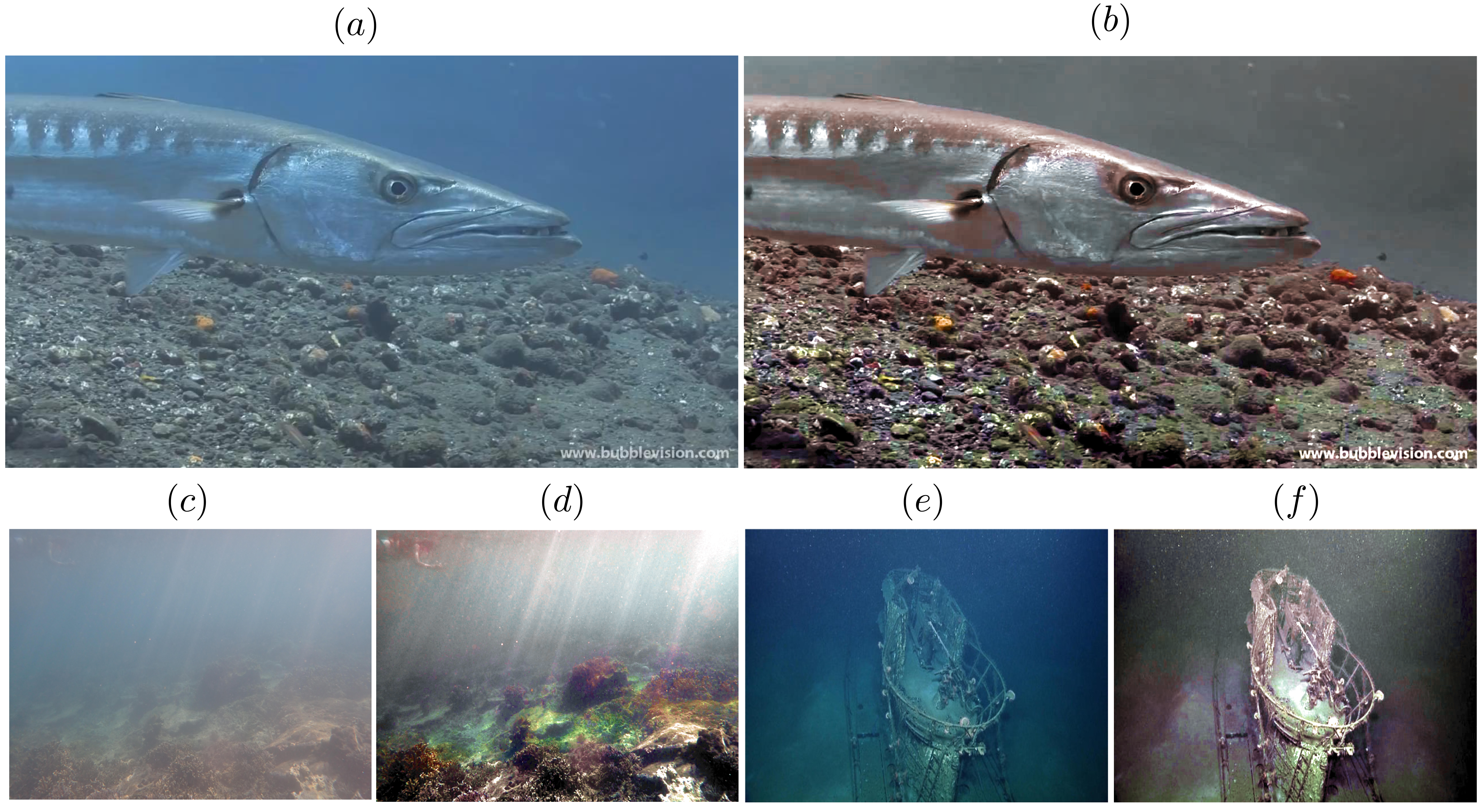}
	\caption{General examples of the proposed method. (a) A typical underwater image showing dominating blue cast, and the contrast is relatively low. (b) Result of the proposed method applied to (a) removes the blue cast and enhances the textures on the riverbed. (c) A blurry underwater image where objects are hardly visible. (d) Result of the proposed method applied to (c) which shows vibrant colors and sharp objects' boundaries. (e) A deep underwater image commonly seen in field exploration. (f) Result of the proposed method applied to (e) which renders the details of the structure of interest.}
	\label{fig_gen_examples}
\end{figure*}

The proposed method  is flexible and adaptive to different image contents. It yields improvements on the image contrasts and color balance which are typically degraded in underwater images.  In Figure~\ref{fig_gen_examples}, we demonstrate various examples where underwater images are used for (a) submarine biology, (c) recreational purposes, and (e) field exploration. Given possible differences in the imaging environment and devices, our method shows consistent and stable behaviors in terms of removing the color casts and enhancing the image quality, and the corresponding processed results are in (b), (d), and (f). 

\subsection{ Different Underwater Color Cast}
\begin{figure}
	\centering
	\includegraphics[width=3.1 in]{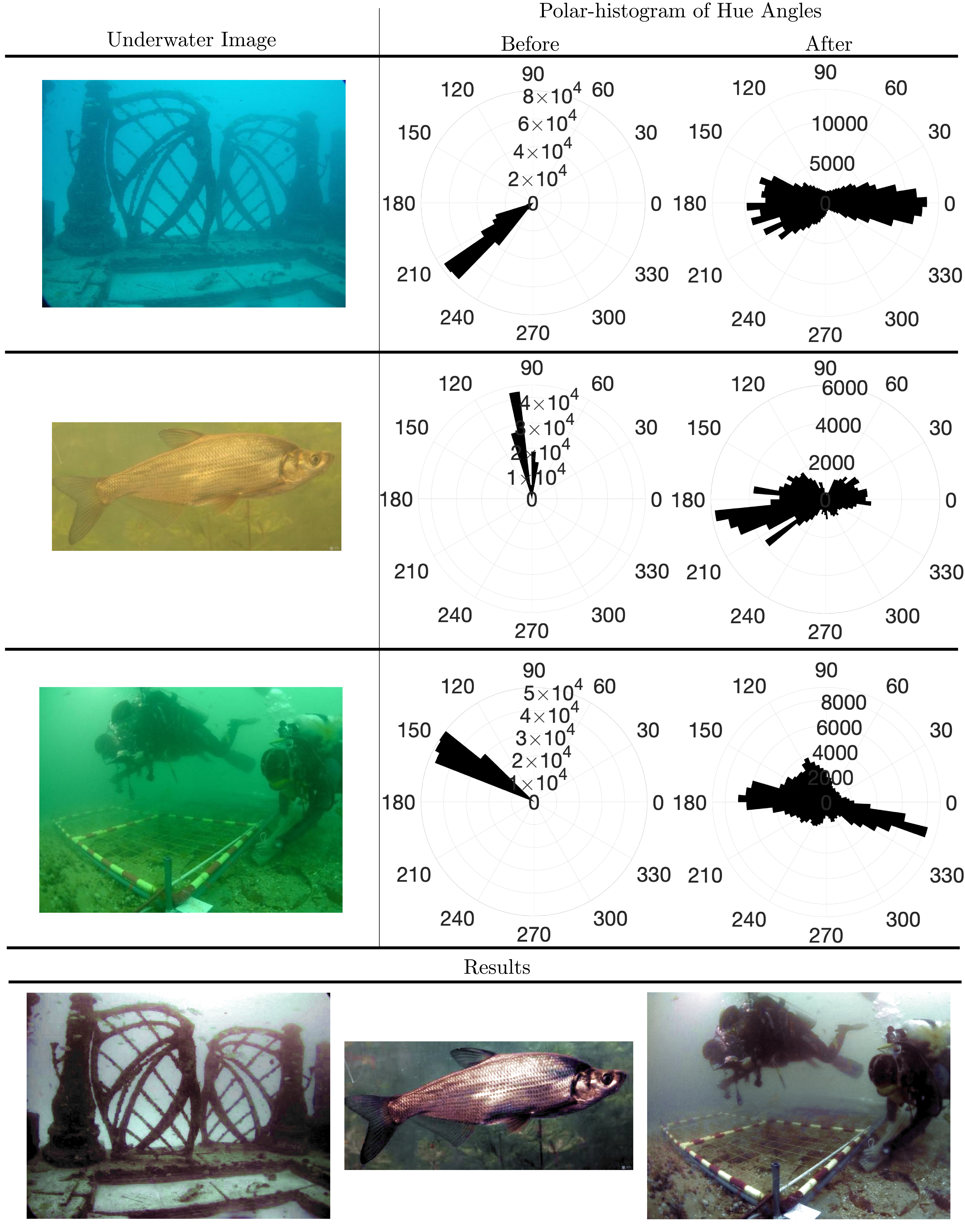}
	\caption{The proposed method shows consistent performance for underwater images with different color casts. For each underwater image in the first 3 rows, we show their hue angle distributions in the second column; in the third columns we show the distribution of the final results, which are displayed in the last row.  In all cases, we have $\eta=6$ and $\beta=1/4$. }\label{fig_adapt_ill}
\end{figure}

Underwater imaging environment is complicated and various conditions can affect the chromatic attributes of the color cast. In Figure~\ref{fig_adapt_ill}, we show the stability of our proposed method for underwater images with different color casts. The scene in the first row shows strong blue veiling light (hue angles concentrating around $210^\circ$), the one in the second row has a yellow color cast (around $95^\circ$), and the third is dominated by a green color (around $150^\circ$). In the third column, we observe that the hue angles of the processed results are more spread out, and the resulted images are displayed in the last row. Although the color casts in the original images are distinct, the colors different from the estimated color casts are preserved and emphasized in the final results.

\subsection{Necessity of the Robust Factor}
The procedure~\eqref{eq_gamma_enhance} enhances the residual colors after the dominating cast is neutralized.  Specifically, the absolute change of the hue angle can be expressed as
\begin{align}
	&|h^\circ_{\text{adapt}}(x,y)-h^\circ_0(x,y)|=\frac{180^\circ}{\pi}\arccos\left(\frac{a_\text{adapt}^*(x,y)a_0^*(x,y)+b_\text{adapt}^*(x,y)b_0^*(x,y)}{C_{\text{adapt}}^*(x,y)C^*_0(x,y)}\right)\;.\label{eq_hue_diff}
\end{align}
Let $C_\mathcal{G}^*(x,y)=\sqrt{(\mathcal{G}_{\sigma_0}*a_0^*(x,y))^2+(\mathcal{G}_{\sigma_0}*b_0^*(x,y))^2}$ be the chroma of the estimated color cast at $(x,y)$, $\rho(x,y) = C_0^*(x,y)/C_\mathcal{G}(x,y)$ as the ratio of image chroma and color cast chroma, and $\gamma(x,y)=(a_0^*(x,y)(\mathcal{G}_\sigma*a^*_0)(x,y)+b_0^*(x,y)(\mathcal{G}_\sigma*b^*_0)(x,y))/(C_0^*(x,y)C_\mathcal{G}^*(x,y))$ as a measure of the hue angle difference between the  image color and the color cast, then~\eqref{eq_hue_diff} can be rewritten as
\begin{align}
	\frac{180^\circ}{\pi}\arccos\left(\frac{\rho(x,y)-\gamma(x,y)}{\sqrt{\rho^2(x,y)-2\gamma(x,y)\rho(x,y)+1}}\right)\;.\label{eq_hue_diff2}
\end{align}
Notice that when the image color and the estimated color cast have similar hue angles, i.e., $\gamma(x,y)\approx1$,~\eqref{eq_hue_diff2} shows that  
\begin{align}
	|h^\circ_{\text{adapt}}(x,y)-h^\circ_0(x,y)|\approx\begin{cases}
		180^\circ\;,\quad&\text{if~}\rho(x,y)<1\\
		90^\circ\;,\quad&\text{if~}\rho(x,y)=1\\
		0^\circ\;,\quad&\text{if~}\rho(x,y)>1
	\end{cases}\;,
\end{align}
which is independent of the lightness.  This implies that a direct enhancement as in~\eqref{eq_gamma_enhance} is very sensitive to the ratio of the image chroma and color cast chroma. 

\begin{figure}
	\centering
	\includegraphics[width=3.1 in]{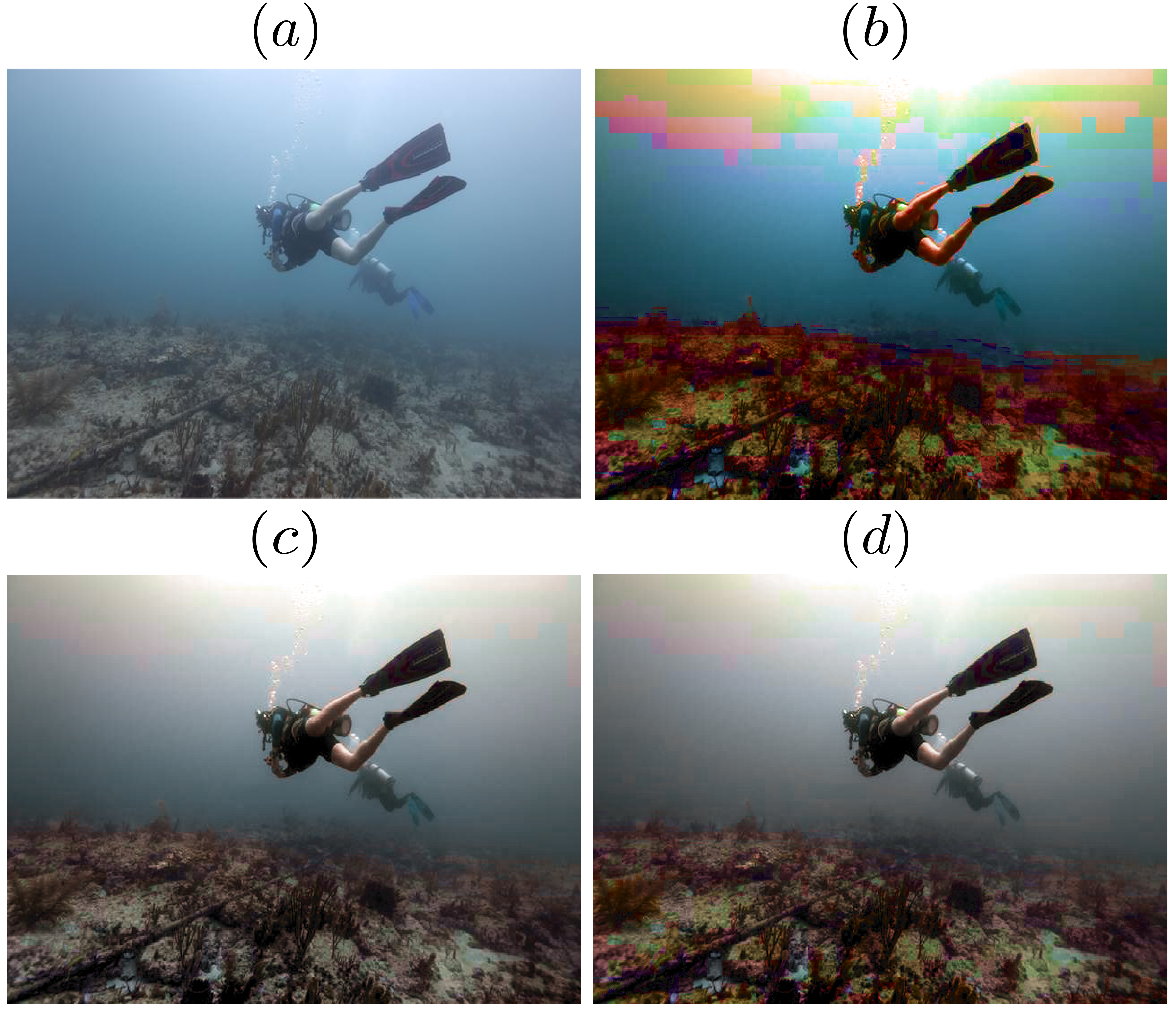}
	\caption{(a) Original underwater image. (b) Proposed method without applying the robust factor~\eqref{eq_zeta} ($\eta=8$). (c) Proposed method without applying the robust factor ($\eta=2$). (d) Proposed method with the robust factor ($\eta=8, \beta = 1/3$). Using the robust factor suppresses the background noisy colors while enhancing the saturation of the other regions.}\label{robust_demo}
\end{figure}

Such instability will cause chromatic noise in areas where the colors are slightly different from the estimated color cast, and typically this happens when the underwater image contains a large portion of background, e.g., Figure~\ref{robust_demo} (a). In Figure~\ref{robust_demo} (b), we show the image resulted from the direct enhancing~\eqref{eq_gamma_enhance} using $\eta=8$, which presents noisy colors in the background region. A possible remedy is to use smaller values of $\eta$, however, this will also subdue the saturation of regions which deserve enhancing. For example, in Figure~\ref{robust_demo} (c) where we use $\eta=2$, although the noisy colors in the background are suppressed, the riverbed becomes almost achromatic.  By using the proposed robust factor~\eqref{eq_zeta}, the corrected image as shown in Figure~\ref{robust_demo} (d) reduces the noise while maintaining a more saturated rendering outside the background region.

\subsection{Behaviors of the Saturation Parameter $\eta$}
\begin{figure}
	\centering
	\includegraphics[width=3.1in]{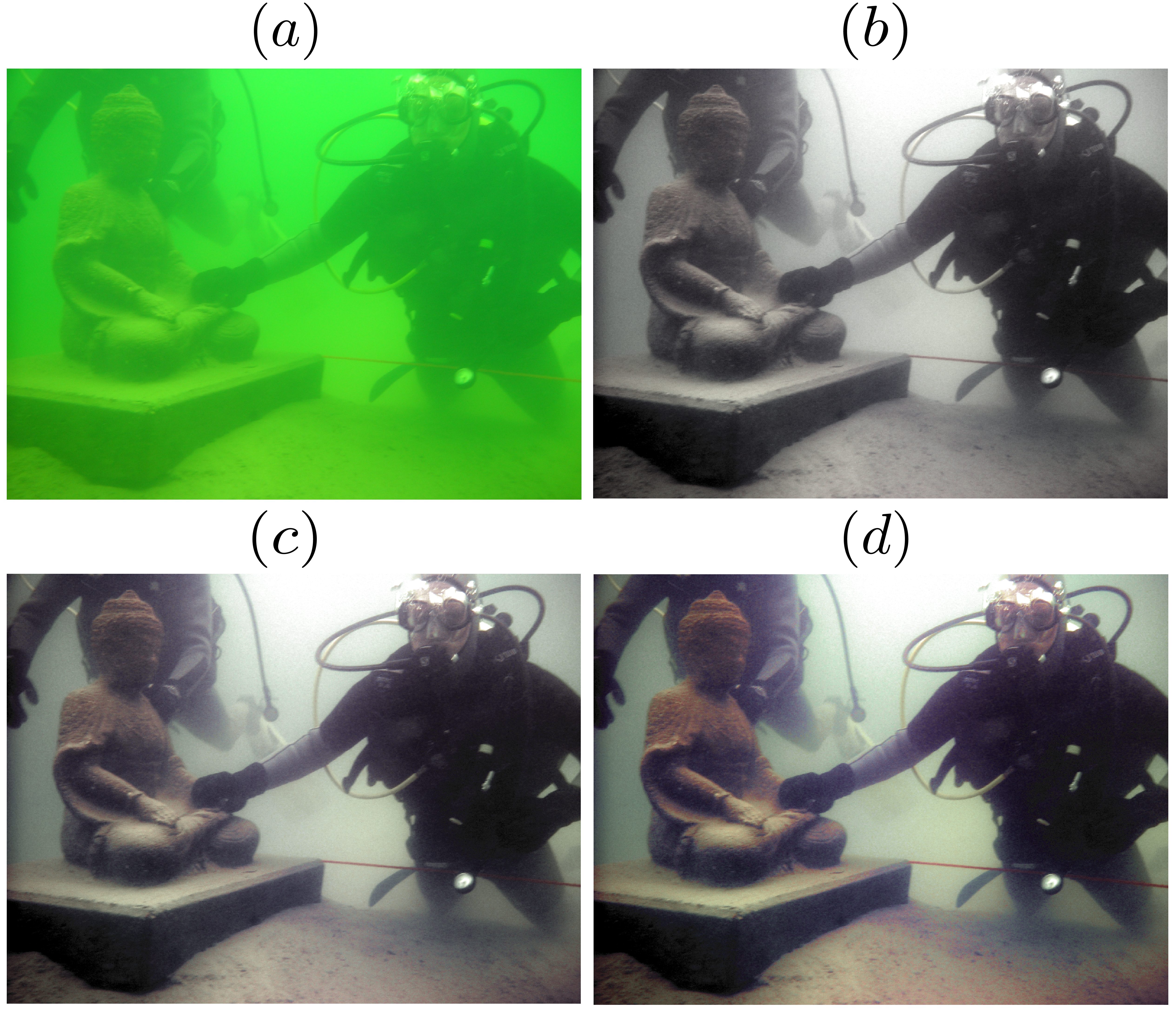}
	\caption{Effect of the chroma enhancement parameter $\eta$. (a) Original underwater image. (b) Result with $\eta=2$. (c) $\eta=4$. (d) $\eta=10$. Here we fix $\beta = 1/4$. }\label{fig_eta}
\end{figure}

The chroma enhancement parameter $\eta$~\eqref{eq_gamma_enhance} allows flexible adjustment of the image saturation. In Figure~\ref{fig_eta}, fixing $\beta=1/3$, we apply the proposed method to the underwater image (a) using $\eta=2$, $\eta=4$ and $\eta=10$, which are shown in (b), (d), and (d), respectively. When we increase the parameter $\eta$, the saturated green color cast in the original image keeps muted, as it is neutralized before the chroma enhancement. As for the objects of colors different from the color cast, e.g., the string and statue, they become more recognizable when greater values of $\eta$ are applied. This example demonstrates two  features of our method. 

First, objects of smaller scales compared to the radius of the Gaussian kernel used in~\eqref{eq_opt_model}  are the most distinguishable in the results.  This is due to the fact that, the  pixels of these objects have little impact on the estimated color cast, thus the complimentary pairs of the objects' colors will not contribute to the neutralization process according to CAT. Consequently, they preserve most of their chromatic properties and get emphasized after the neutralization of the color cast and the chroma enhancement. For instance, see the red string, the white oxymeter, and the texture of the sand. 

Second, although enhancing the saturation by~\eqref{eq_gamma_enhance} is global, the background color remain relatively muted compared to others during this process. When we increase the value of $\eta$, the red string and brown statue become more saturated than the green background. This can render the objects in the scene more distinguishable and help improve the image contrast in a global scale. 

\subsection{Qulitative Comparison}
\begin{figure*}
	\centering
	\begin{tabular}{ccc}
		(a) Original Image&(b) Zhao et al.&(c) Peng et al.\\
		\includegraphics[width=0.3\textwidth]{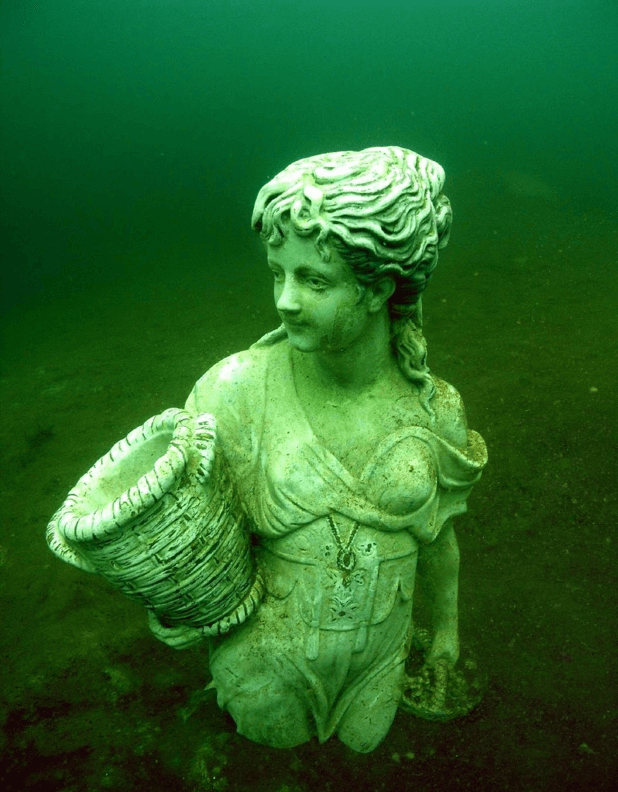}&
		\includegraphics[width=0.3\textwidth]{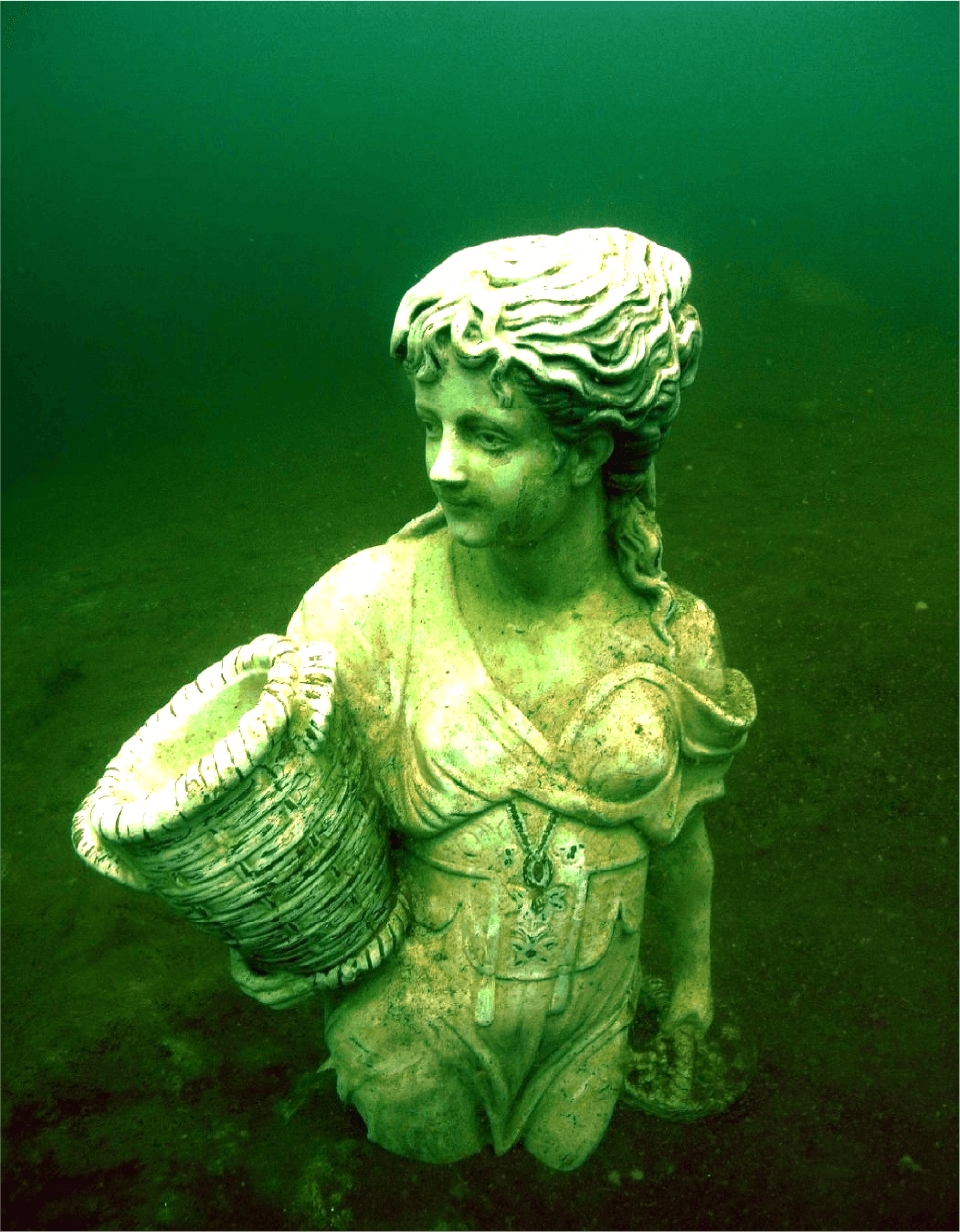}&
		\includegraphics[width=0.3\textwidth]{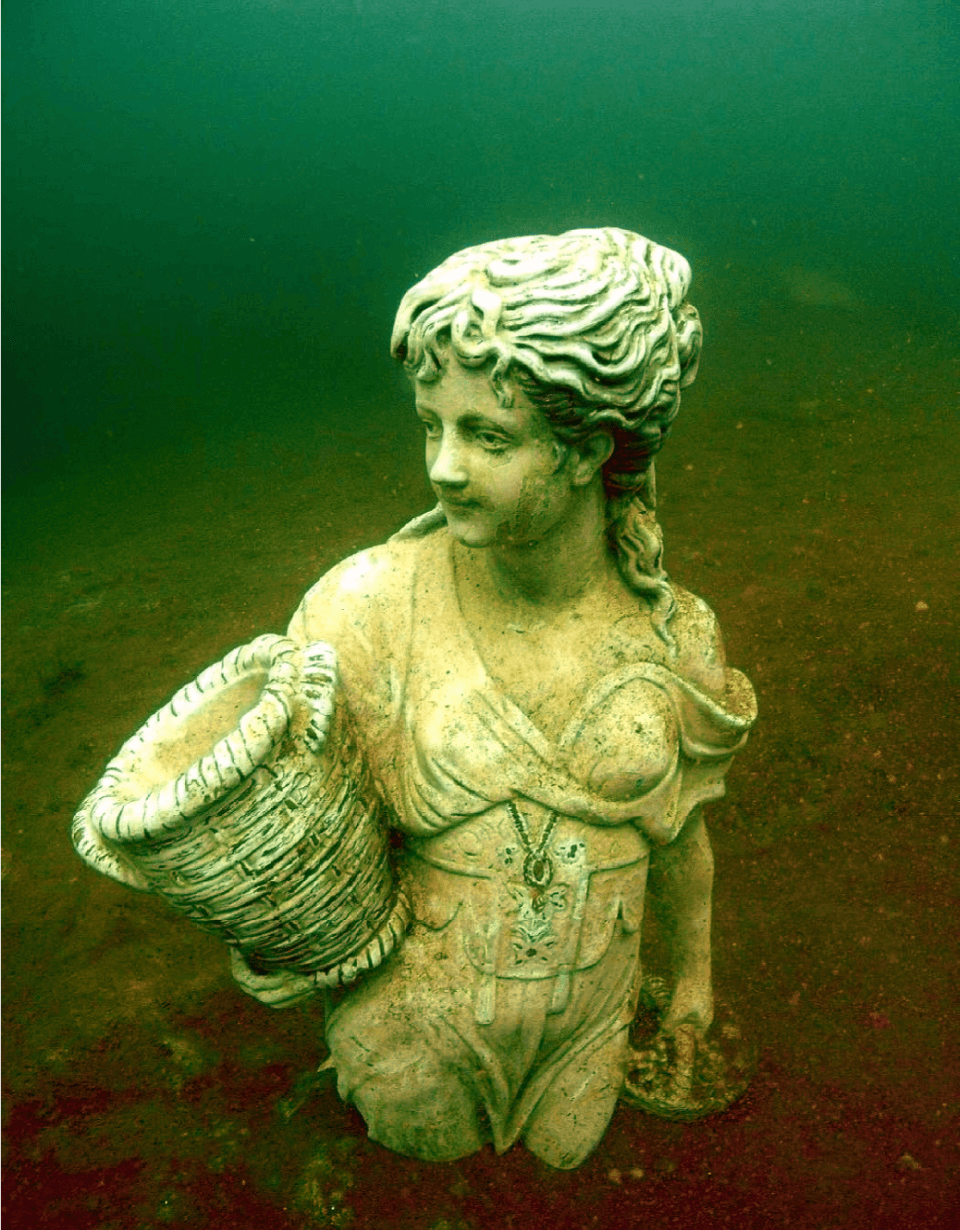}\\
		(d) Histogram Equalization&(e) Limare et al. &(f) Automatic Color Enhancement\\
		\includegraphics[width=0.3\textwidth]{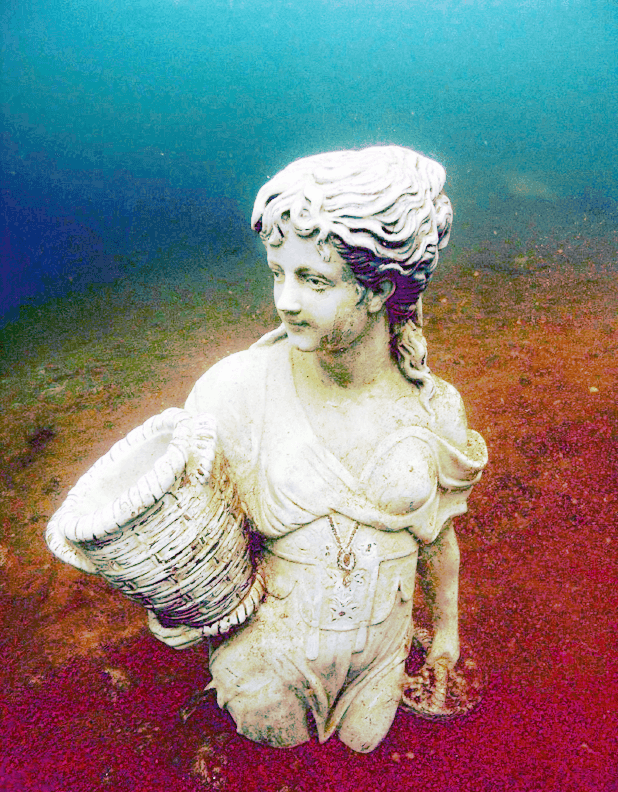}&
		\includegraphics[width=0.3\textwidth]{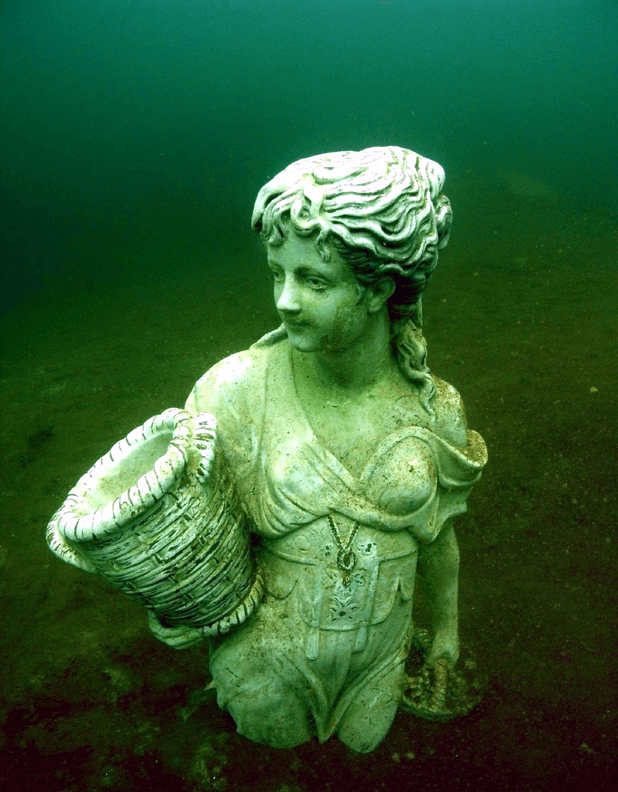}&
		\includegraphics[width=0.3\textwidth]{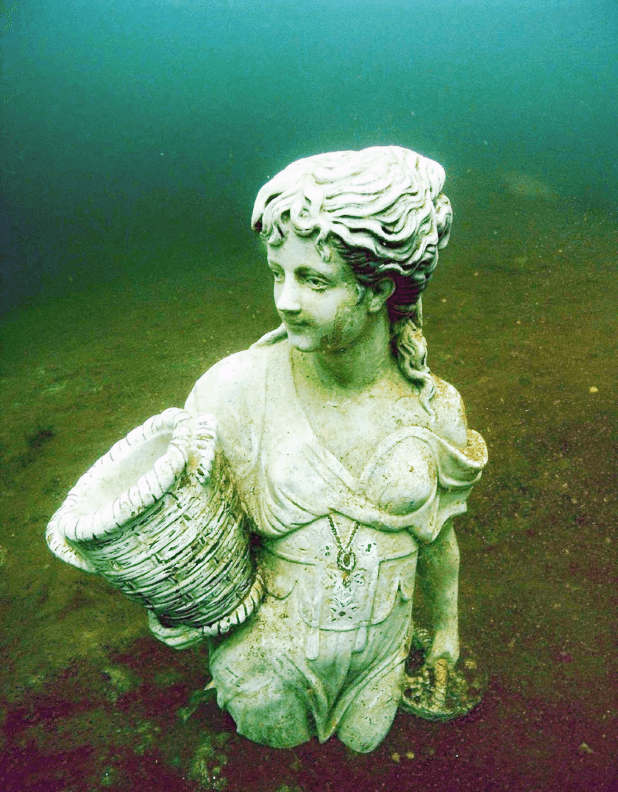}\\
		(g) Local Color Correction&(h) Multiscale Retinex &(i) Proposed Method\\
		\includegraphics[width=0.3\textwidth]{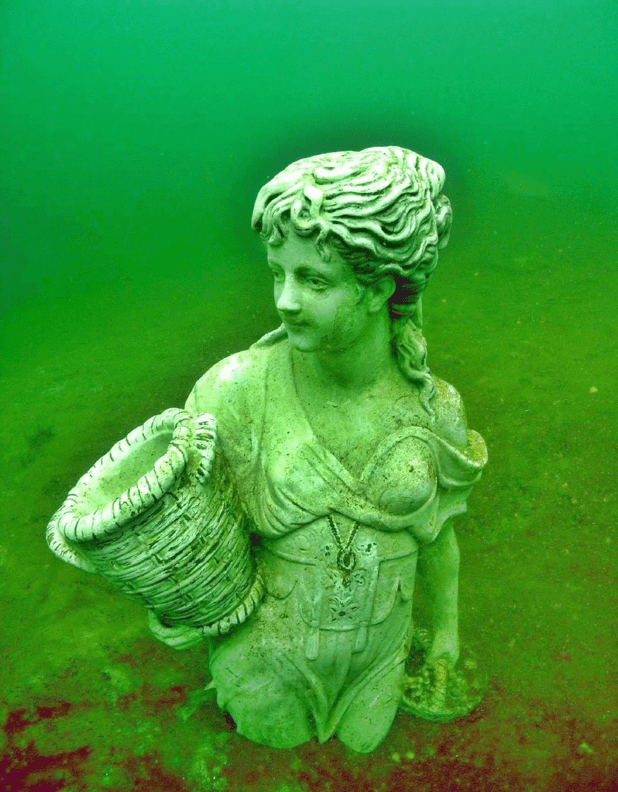}&
		\includegraphics[width=0.3\textwidth]{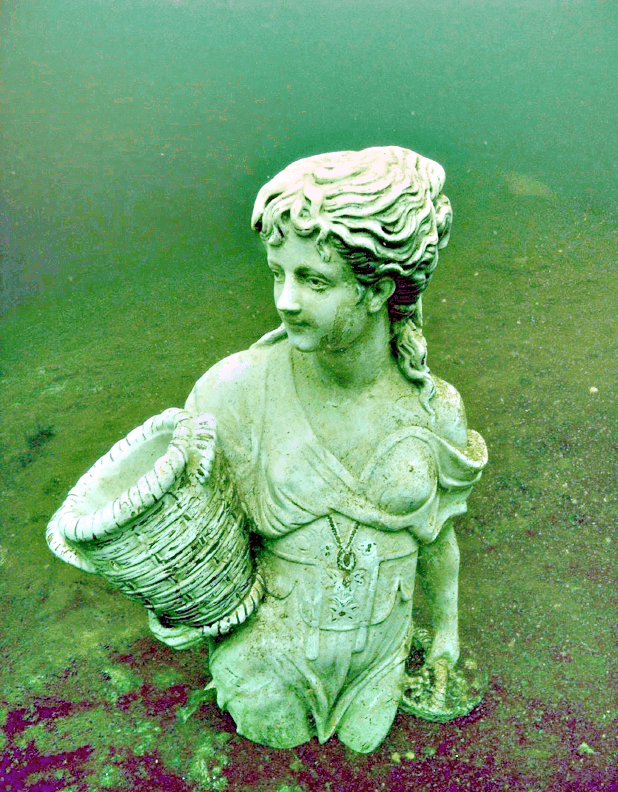}&
		\includegraphics[width=0.3\textwidth]{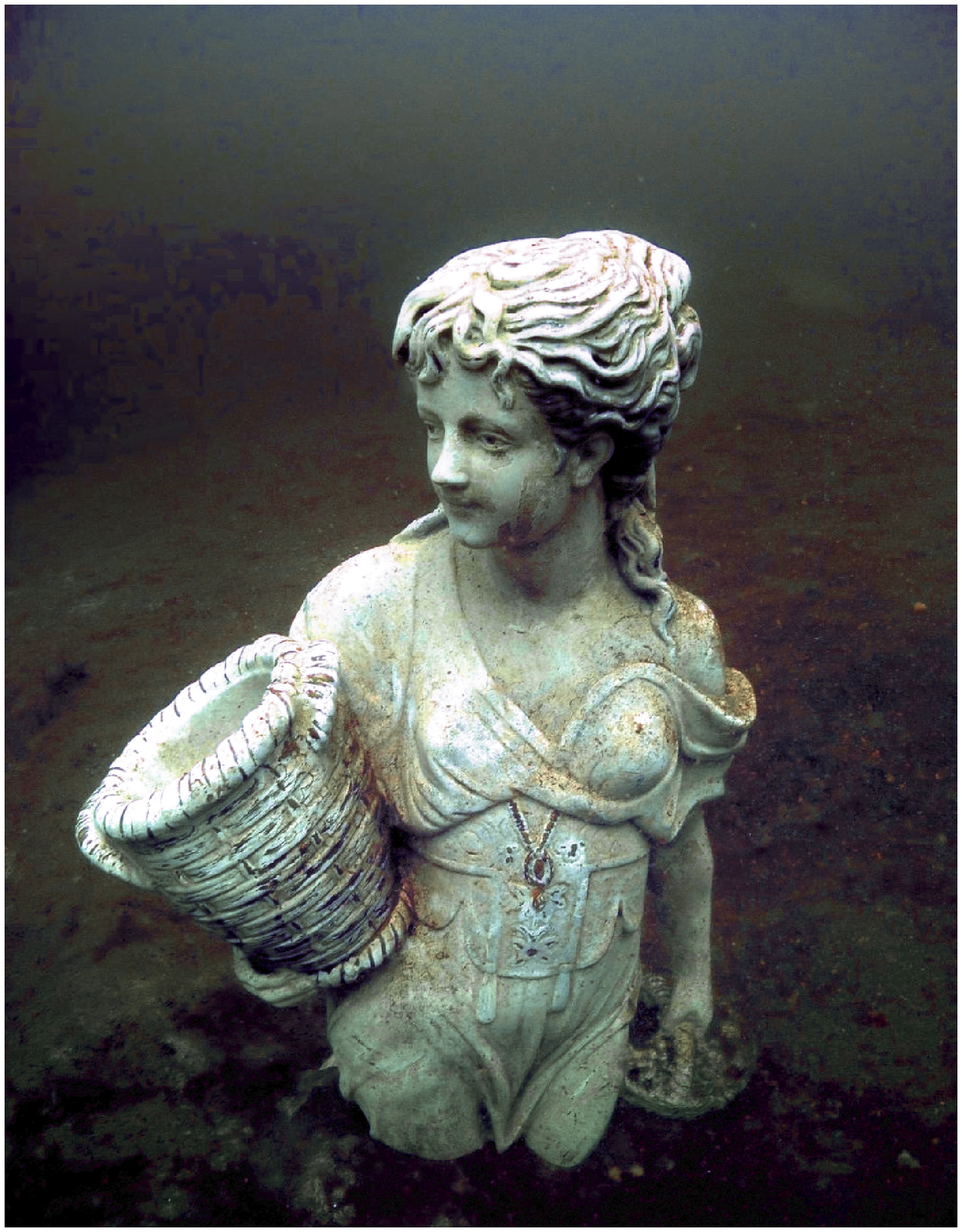}
	\end{tabular}
	\caption{Qualitative comparison of different methods for underwater images. (a) Original underwater image. (b) Zhao et al.~\cite{ZhaoDeriving} (c) Peng et al.~\cite{peng2017underwater} (d) Histogram Equalization (HE) (e) Limare et al.~\cite{ipol.2011.llmps-scb} (f) Automatic Color Enhancement (ACE)~\cite{bertalmio2007perceptual,ipol.2012.g-ace} (g)Local Color Correction~\cite{ipol.2011.gl_lcc} (h) Multiscale Retinex~\cite{rahman1996multi,ipol.2014.107} (i) Proposed Method. Our method has distinct characteristics compared to the others: effective color cast removal, prominent enhancement on the colors of small-scale textures, and sharp objects' boundaries.}\label{fig_qual}
\end{figure*}

We compare our proposed method to some of the state-of-art approaches in the literature. They are designed either specifically for underwater images, or for color constancy for general color images. On the oirginal image in Figure~\ref{fig_qual} (a), we compare Zhao et al.~\cite{ZhaoDeriving} (shown in (b)), Peng et al.~\cite{peng2017underwater} (shown in (c)), Histogram Equalization (shown in (d)), Limare et al.~\cite{ipol.2011.llmps-scb} (shown in (e)),  Automatic Color Correction (ACE)~\cite{bertalmio2007perceptual,ipol.2012.g-ace}, Local Color Correction~\cite{ipol.2011.gl_lcc} (shown in (g)),  Multiscale Retinex~\cite{rahman1996multi,ipol.2014.107} (shown in (h)), and the proposed method in (i). We see that these methods exhibit different chromatic properties in their results.

Both (b) and (c) are obtained from underwater-image-specific approaches based on the Koschmieder model. They differ from each other by the techniques used for background light and transmission map estimation. As pointed out in~\cite{peng2017underwater}, these estimated quantities determine the results, and in many cases, such relation is very sensitive. With careful combination of different priors and estimations, the result in (c) shows better color restoration on some region of the statue and the riverbed compared to (b).

The methods used in the second row manipulate the image histogram. For (d), we see the typical over-saturation in HE. The method in (e), which aims at enhancing the dynamical ranges of the RGB-channels, does not show effective color balancing in this example. As a localized version of HE, (f) renders more realistic colors compared to (d). 

The method for (g) is based on a nonlinear filter applied in the HSL color space, which demonstrate enhancement on the brightness and saturation, yet the green cast is not removed. The Multiscale Retinex used in (h) improves the image brightness and makes many textures visible, but the colors are still biased toward green. 

The proposed method in (i) shows distinct visual perception from the others. First, the strong green cast in the original underwater image is effectively removed. This renders a neutral background and recovers realistic tones for the statue, which is perceived as white in (a). Second, the colors for small-scale textures become apparent. For example, we clearly see the brownish mud around the statue and the color variations on the riverbed. Third, because of the neutralization of the background and the enhancement on the small scale contents, our result shows better contrast improvement. Notice the head and shoulder of the statue, as well as the patterns on the pottery.

\subsection{Quantitative Evaluation and Comparison}
In this set of experiments, we evaluate the performances of the proposed algorithm and other methods in the literature using two measures:  Underwater Color Image Quality Evaluation metric (UCIQE)~\cite{yang2015underwater} and Underwater Image Quality Measure (UIQM)~\cite{panetta2016human}. These metrics are specifically designed to evaluate the  quality of underwater images.  Both  UCIQE and UIQM are linear combinations of certain image attributes such as colorfulness,  saturation, and contrast, whose coefficients are statistically derived. Higher values of these metrics indicate better image qualities.

\begin{figure*}
	\centering
	\includegraphics[width=0.9\textwidth]{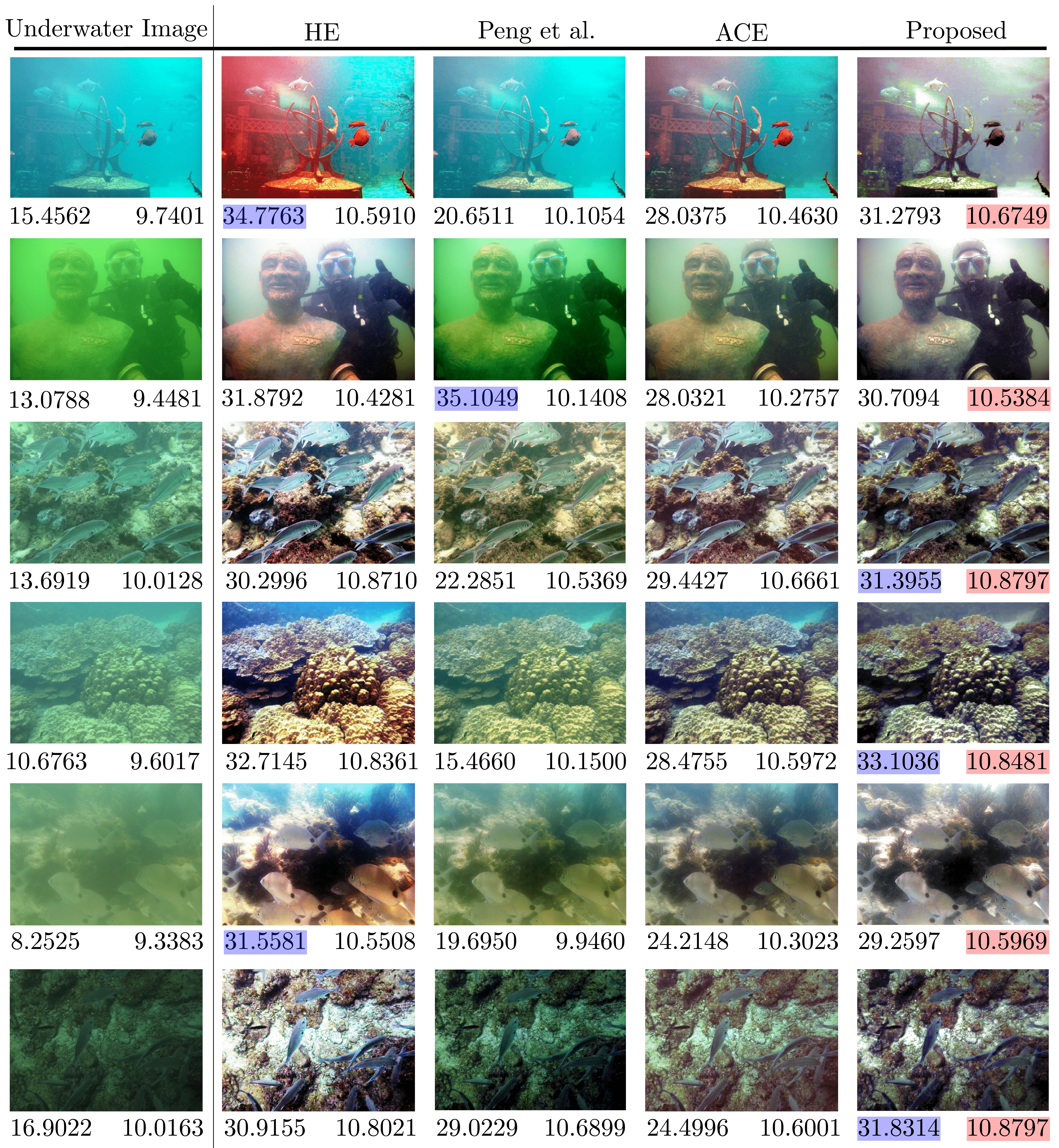}
	\caption{Quantitative evaluation and comparison. We compare our methods with Histogram Equalization (HE), Peng et al.~\cite{peng2017underwater}, and Automatic Color Enhancement (ACE)~\cite{bertalmio2007perceptual,ipol.2012.g-ace}. The quality of each image is evaluated by UCIQE~\cite{yang2015underwater} (left, blue marks the best) and UIQM~\cite{panetta2016human} (right, red marks the best). In all the cases, we use $\eta=10$ and $\beta=1/4$ for the proposed method.}
	\label{fig_quant}
\end{figure*}

Figure~\ref{fig_quant} collectively shows the results from Histogram Equalization, Peng et al.~\cite{peng2017underwater}, Automatic Color Enhancement (ACE)~\cite{bertalmio2007perceptual}, and the proposed method, where the underwater images have various contents and complexities.  Our method performs consistently the best measured by UIQM, and the values of UICQE for some of our results are the highest. Among all the methods in comparison, HE produces the most colorful results,  yet some of which are overly saturated. The method proposed by Peng et al. performs well when the veiling light color is correctly estimated. ACE is a local HE in principle, hence we observe similar chromatic features between them. Compared to HE, ACE produces more natural color distributions.  Among the results from the proposed method, observe that a common characteristic is that the strong color casts in the original underwater images are neutralized. This feature induces a visual effect that the objects  against the original saturated background have sharper boundaries. 

\section{Conclusion}\label{sec_conclude}

In this paper, we presented a new mathematical  interpretation of the complimentary adaptation theory proposed by Gibson in 1937. As an alternative of the well-known Retinex theory for understanding the color constancy, CAT emphasizes the neutralization function of the complementary pair of the lasting stimulus rather than HVS's competence of identifying the reflectance. We modeled this adaptation process as a Tikhonov-type optimization problem in the CIELAB color space. This is a simple model which produces the adapted colors as a balance between the original underwater image and the complimentary pair of the estimated color cast. We overcame the lack of uniformity of CIELAB by employing two techniques: a pre-processing compensating the hue-distortion in the blue region, and a post-processing addressing the H-K effect. Numerically, we demonstrated the necessities of the introduced techniques and qualitatively compared our model with some of the state-of-art methods for underwater images. The proposed method shows superior stability when dealing with various underwater environments and recovers realistic colors compatible with the visual perception.

\appendix

\section{CIELAB  Boundary Estimation}\label{sec_gamut_boundary}
\begin{figure}
	\centering
	\includegraphics[width=3.1in]{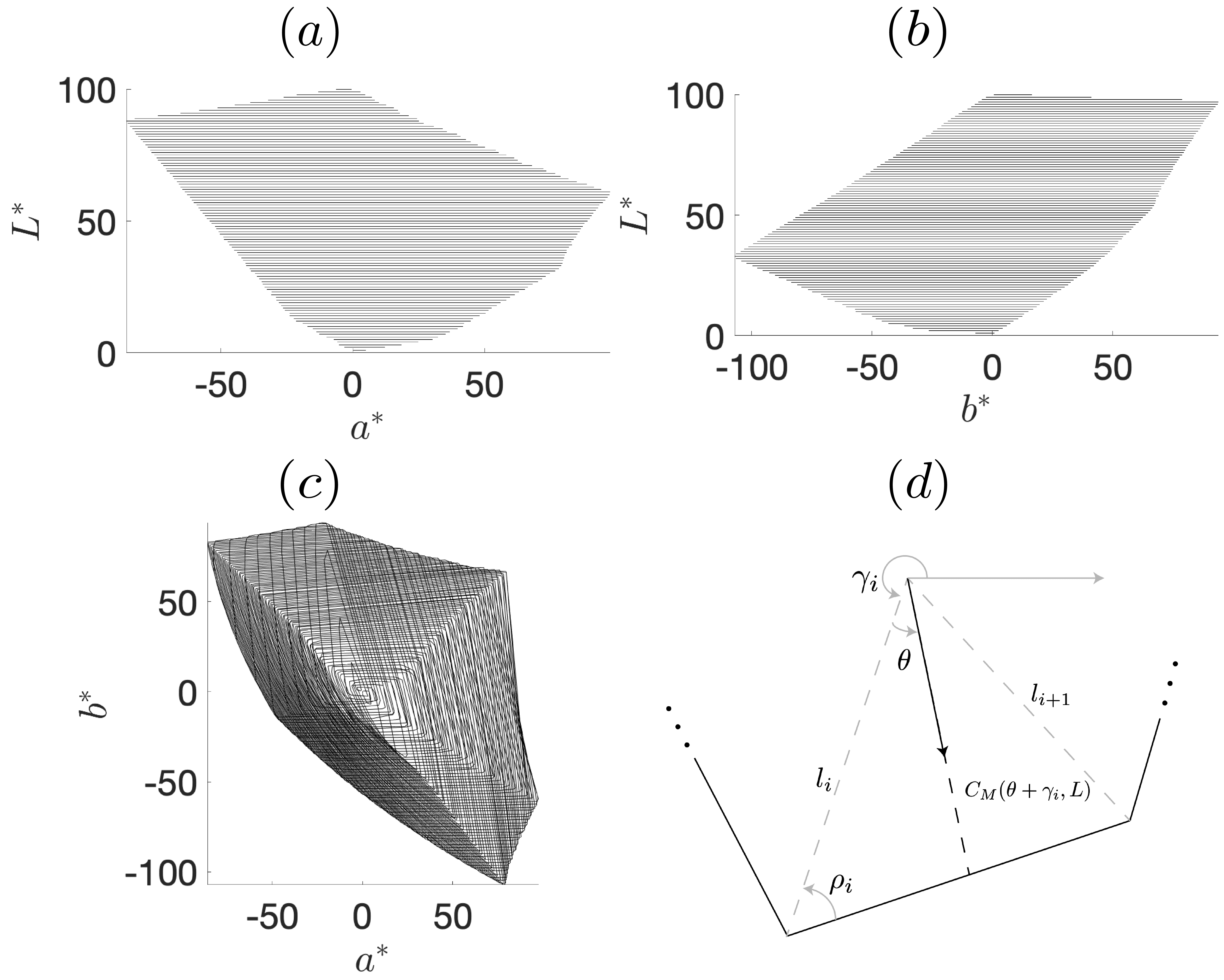}
	\caption{(a) CIELAB gamut projection on the $L^*$-$a^*$ plane. (b) Projection on the $L^*$-$b^*$ plane. (c) Projection on the $a^*$-$b^*$ plane. (d) Geometry for computing the chroma upper limit $C^*_{\text{max}}(L,\theta+\gamma_i)$ in the direction of the hue angle $\theta+\gamma_i$ on the lightness level of $L$. Here the chroma direction falls within the sector $[\gamma_{i},\gamma_{i+1}]$, $i=1,2,\dots$.}\label{fig_gamut}
\end{figure}

The RGB color space is geometrically a cube embedded in the Euclidean space $\mathbb{R}^3$, however, the transformation from RGB to CIELAB maps the RGB cube to an irregular shape. See the illustration in Figure~\ref{fig_gamut} (a)--(c). Here, we randomly sample $500000$ points in the RGB cube, convert them to the CIELAB space, digitize their lightness coordinates by taking the ceil function, and compute the convex hull of the chromaticity section for each digitized lightness. Hence, the CIELAB gamut is approximated by the union of these convex slices at different lightness levels
\begin{align}
	\bigcup_{L^*=0}^{100}\text{conv}\{V_i(L^*)\}_{i=1}^{N(L^*)}\;,
\end{align}
where $V_i(L^*)$ represents a vertex of the convex hull computed using the color samples with digitized lightness $L^*$, and $\text{conv}\{\cdot\}$ computes the convex hull supported by a finite set of points.  This approach offers a simple estimation of the chroma boundary given the lightness $L^*$ and hue angle $h^\circ$ according to basic trigonometry
\begin{align}
	&C^*_{\max}(L^*,h^\circ)=\frac{l_{j(h^\circ)}\sin(\rho_{j(h^\circ)})}{\sin(\pi-\theta_{j(h^\circ)}-\rho_{j(h^\circ)})}\;,\nonumber\\
	& j(h^\circ)\in\{1,2,\dots,N(L^*)\}\label{eq_Cmax}
\end{align}
where $\theta_{j(h^\circ)}<h^\circ<\theta_{j(h^\circ)+1}$, $\theta_{j(h^\circ)}$ is the angle from the positive direction of the  $a^*$-axis to $V_{j(h^\circ)}(L^*)$ in a counter-clockwise orientation, $\theta_{N(L^*)+1}$ takes $\theta_{1}$, and $l_{j(h^\circ)}$ is the distance from $V_{j(h^\circ)}(L^*)$ to the origin. Both quantities $l_{j(h^\circ)}$ and $\theta_{j(h^\circ)}$ depend on $L^*$, and we suppress this notation in~\eqref{eq_Cmax} for simplicity. See Figure~\ref{fig_gamut} (d) for an illustration.

\section*{Acknowledgment}

The author would like to thank Professor Sung Ha Kang from School of Mathematics of Georgia Institute of Technology for valuable discussion and suggestions, and Dr. Chongyi Li from City University of Hong Kong for sharing the data set of underwater images used in this paper.

	\bibliographystyle{plain}
	\bibliography{underwater_bib}
\end{document}